\documentclass[letterpaper, 10 pt, conference]{ieeeconf}  % Comment this line out if you need a4paper

\IEEEoverridecommandlockouts                              % This command is only needed if 
\let\labelindent\relax 
\usepackage[inline]{enumitem}
\usepackage[utf8]{inputenc} % allow utf-8 input
\usepackage[T1]{fontenc}    % use 8-bit T1 fonts
\usepackage{hyperref}       % hyperlinks
\usepackage{footmisc}
\usepackage{url}            % simple URL typesetting
\usepackage{booktabs}       % professional-quality tables
\usepackage{amsfonts}       % blackboard math symbols
\usepackage{nicefrac}       % compact symbols for 1/2, etc.
\usepackage{microtype}      % microtypography
\usepackage{xcolor}         % colors
\usepackage[table]{xcolor}
\usepackage{multirow}
\usepackage{caption}
\usepackage{colortbl}
\usepackage{makecell}
\usepackage{helvet}
\usepackage{bm}
\usepackage{amsmath}
\usepackage{mdframed}
\usepackage{amssymb}
\usepackage{algorithm}    
\usepackage{algpseudocode} 
\usepackage{graphicx}%
\usepackage{wrapfig}  % 实现文字环绕

\begin{document}

\title{\LARGE \bf InSpire: Vision-Language-Action  Models  with Intrinsic Spatial Reasoning }
% \author{Anonymous authors}

\author{%
  Ji Zhang$^{1*}$\thanks{* Equal contribution} \quad \quad  \quad Shihan Wu$^{2*}$   \quad \quad \quad Xu Luo$^2$ \quad \quad \quad \quad Hao Wu$^2$ \\
  \textbf{Lianli Gao}$^2$ \quad \quad  \quad \textbf{Heng Tao Shen}$^3$ \quad  \quad \quad \textbf{Jingkuan Song}$^{3\dag}$\thanks{\dag\ Corresponding author}  \\
  $^1$Southwest Jiaotong University\\
  $^2$University of Electronic Science and Technology of China  \\ $^3$Tongji University  \\ \texttt{\{jizhang.jim,jingkuan.song\}@gmail.com} 
    \\ \\
  \textcolor{magenta}{\url{https://koorye.github.io/proj/Inspire}}
}

\maketitle

\begin{abstract}
Leveraging pretrained Vision-Language Models (VLMs) to map language instruction and visual observations to raw low-level actions, Vision-Language-Action models (VLAs) hold great promise for achieving general-purpose robotic systems. 
% Vision-Language-Action models (VLAs), trained end-to-end for outputting raw low-level actions in response to visual % observations, hold great promise for achieving general-purpose and user-friendly robotic systems. 
Despite their advancements, existing VLAs tend to spuriously correlate task-irrelevant visual features with actions, limiting their generalization capacity beyond the training data.
To tackle this challenge, we propose \textbf{Intrinsic Spatial Reasoning (InSpire)}, a simple yet effective approach that mitigates the adverse effects of spurious correlations by boosting the spatial reasoning ability of VLAs.
% During training, our InSpire redirects the model's attention to task-specific visual tokens by simply appending the question “\textit{In which direction is the [object] relative to the robot}” before the language instruction and aligning the output tokens preceding the actions to the ground-truth answer “\textit{right / left / up / down / grasp}”. 
Specifically, InSpire redirects the VLA's attention to task-relevant factors by prepending the question “$\mathsf{In\,\, which \,\, direction\,\, is\,\, the\,\, [object]\,\, relative\,\, to\,\, the\,\, robot ?}$” to the language instruction and aligning the model's output answer “$\mathsf{right/left/up/down/front/back/grasped}$” and predicted actions with ground-truth.
% performing action prediction and object-oriented spatial reasoning at the same time.
% visual question answering task.
Notably, InSpire can be used as a \textit{plugin} to enhance existing autoregressive VLAs, requiring no extra training data or interaction with other large models.
Extensive experimental results in both simulation and real-world environments demonstrate the effectiveness and flexibility of InSpire. 
% InSpire improves the success rates of state-of-the-art VLAs by \textbf{10}\%$\sim$ \textbf{20}\% in real-world manipulation tasks and \textbf{30}\%$\sim$\textbf{40}\% in simulation benchmarks.
% with a \textbf{7}$\boldsymbol{\times}$ smaller model.
% Code, pretrained models and demos are publicly available at: \url{https://github.com/InspireVLA/Inspire}.
\end{abstract}

% \keywords{Robotic Foundation Models, Policy Contrastive Decoding } 

%===============================================================================

\section{Introduction}
% Recent advancements in robot learning have made substantial progress in training Vision-Language-Action models (VLAs) to achieve flexible, general-purpose, and dexterous robotic systems.
In recent years, Vision-Language Models (VLMs) \cite{siglip2023,llama2023} have demonstrated remarkable abilities across a diverse set of tasks, including image captioning and visual question answering (VQA). These advances have paved the way for Vision-Language-Action (VLA) models, which use pretrained VLMs to directly map language commands and visual inputs to low-level motor actions, offering a promising pathway toward general-purpose robotic systems \cite{brohan2022rt1,kim2024openvla,team2024octo}.
% Vision-Language-Action models have significantly advanced the development of flexible, general-purpose, and dexterous robotic systems.
% Vision language models (VLMs) have made significant progress in recent years across a variety of tasks including image captioning, visual question answering (VQA), and grounding.
% demonstrate their ability to execute a broad array of tasks with minimal human supervision.
% for achieving general-purpose and user-friendly robotic systems. 
% Specifically, they take as input a visual observation of the robot’s state combined with a language instruction that defines the task, and output a robot action, such as end-effector displacement. 
% Ongoing advancements in scaling real-world robotic data corpora have facilitated the development of numerous VLAs \cite{brohan2022rt1,kim2024openvla,team2024octo,pertsch2025fast}, which have demonstrated exceptional effectiveness in controlling various robots across diverse environments and acquiring a wide range of manipulation skills.
% The scaling up of real-world robotic data corpora has facilitated the development of numerous VLAs that can understand diverse scenes, objects, and natural language instructions \cite{brohan2022rt1,kim2024openvla,team2024octo,pertsch2025fast}.

\begin{figure}[t] 
	\centering
	\includegraphics[width=1\linewidth]{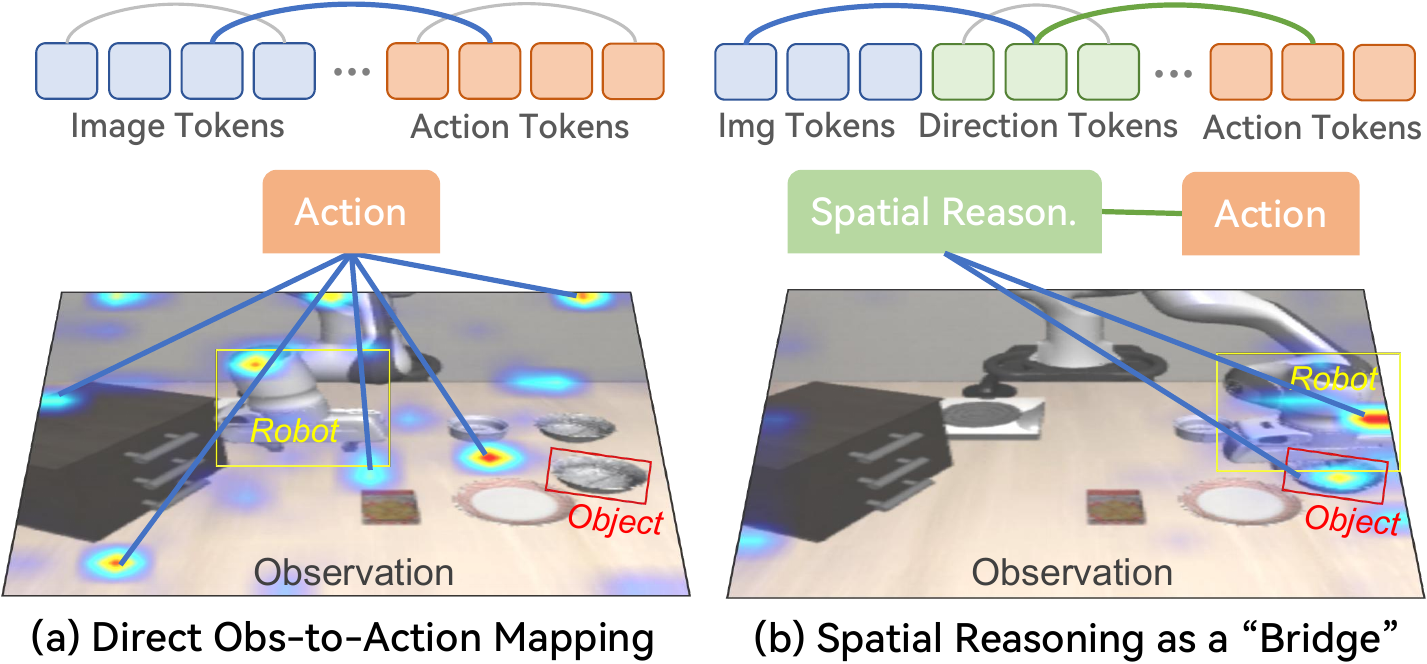} 
	\caption{
    \textbf{(a)} VLAs typically predict actions relying on \textit{Spurious Correlations} learned by the direct observation-to-action mapping mechanism.
    \textbf{(b)} The core idea of our proposed InSpire method that tackles spurious correlations by boosting the spatial reasoning capabilities of VLAs.} \label{mtv}
% \end{figure*}
\label{fig.mtv}
\end{figure}

Despite their advancements, state-of-the-art VLAs tend to spuriously correlate task-irrelevant visual features with actions, overlooking vital elements such as language instructions and spatial relations in visual observations, which hinders their ability to generalize beyond the training data distribution.
% compromises their robustness and generalization in unseen scenarios.
% This leads to a failure in capturing the true causal relationships between observations and actions. Consequently, models may overlook essential elements like language instructions and target objects, thereby restricting their ability to generalize beyond the training distribution.  (i.e., the “black bowl” and “plate”)
As shown in Fig. \ref{fig.mtv} \textbf{(a)}, the VLA—trained via direct observation-to-action mapping—fails to accurately identify task-specific objects and model their spatial relationships with the robot.
Instead, it allocates more attention to task-irrelevant regions when predicting actions.
% Yet, one of the most exciting capabilities of large vision-language models in other domains is their ability to reason iteratively through complex problems.
% This suggests that VLAs, learned through direct observation-to-action mapping mechanisms, are unable to mitigate the VLAs' (or the embedded VLMs') limitations in spatial reasoning \cite{wang2024picture,liu2023visual,chen2024spatialvlm,zhang2024spatialvla}.
% improve the embedded VLM's natural capacity for reasoning about spatial relationships.
% Understanding and reasoning about spatial relationships is a fundamental capability for Visual Question Answering (VQA) and robotics In contrast, humans tend to think more carefully before acting: failing to capture the true causal relationships between observations and actions.
Understanding and reasoning about spatial relationships are crucial skills that empower humans to solve complex tasks: if asked the same question, they would first try to assess the directions (or coarse locations) of the “black bowl” and the “plate” relative to their hands.
% To capture the true causal relationships between observations and actions.
% In order to address the spurious correlation issue in VLAs, it is of great importance to 
% In the same way, we would like our VLAs to perform 
% Motivated by the .
This underscores the importance of leveraging spatial reasoning as a bridge to capture the causal relationships between observations and actions, thereby allowing VLAs to produce more
accurate and robust robot actions, as illustrated in Fig. \ref{fig.mtv} \textbf{(b)}.
Prior efforts generally resort to auxiliary training data \cite{zhao2025cot} or other large models \cite{michal2024robotic,myers2024policy} to enhance chain-of-thought (or step-by-step) reasoning capabilities. 
While these approaches offer partial improvements in spatial reasoning, they often suffer from limitations in efficiency and generalizability.
% Current solutions to this problem generally rely on constructing large auxiliary datasets or leveraging existing large models to explore step-by-step reasoning capabilities. While these approaches have shown some success, they often lack efficiency, flexibility, and generalizability.
% This indicates the need for a spatial reasoning step to \textit{bridge} visual observations with raw low-level actions to address the spurious correlation issue in VLAs (as shown in Fig. \ref{fig.mtv} \textbf{(b)}).
We thus raise the following question: 
% Hence, a research question remains open in the field of robot learning: 
% Thus, we raise the following question:
\begin{mdframed}
[skipabove=4pt, innertopmargin=4pt, innerbottommargin=4pt]
\textit{Without relying on extra data or interacting with other large models, can we enhance the spatial reasoning capabilities of VLAs to resolve spurious correlations?}
\end{mdframed}
% \vspace{-4pt}

% \begin{figure*}[t] 
% % \setlength{\abovecaptionskip}{0.1cm}  x
% % \setlength{\belowcaptionskip}{-0.3cm} 
% 	\centering
% 	\includegraphics[width=1\linewidth]{figs/motivation7.pdf} 
% 	\caption{
%     \textbf{(a)} VLAs typically predict actions relying on \textit{Spurious Correlations} learned by the direct observation-to-action mapping mechanism.  
%     \textbf{(b)} The core idea of our InSpire method that tackles spurious correlations by boosting the spatial reasoning capabilities of VLAs. \textbf{(c)} InSpire can be used as a \textit{plugin} to improve state-of-the-art VLAs on both seen and unseen tasks across simulation and real-world environments. For illustrative purposes, we omit the tokens of the language instruction “{pick up the black bowl next to the plate and place it on the plate}”.} \label{mtv}
% % \end{figure*}
% \label{fig.mtv}
% \end{figure*}

\begin{figure*}[t] 
	\centering
	\includegraphics[width=0.8\linewidth]{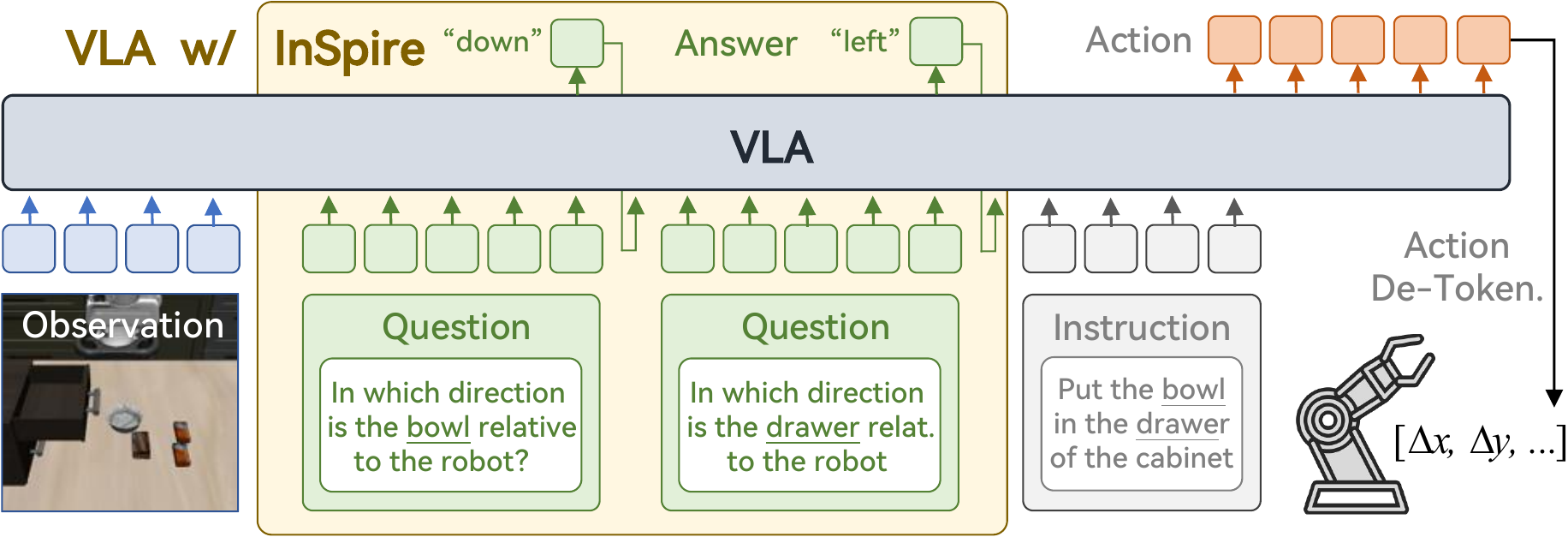} 
	\caption{\textbf{Overview of our InSpire approach}. 
    % During training, our InSpire redirects the model's attention to task-specific visual tokens by simply appending the question “\textit{In which direction is the [object] relative to the robot}” before the language instruction and aligning the output tokens preceding the actions to the ground-truth answer “\textit{right / left / up / down / grasped}”. 
    InSpire boosts the VLA's spatial reasoning ability by appending the question “$\mathsf{In\,\, which \,\, direction\,\, is\,\, the\,\, [object]\,\, relative\,\, to\,\, the\,\, robot ?}$” before the language instruction and aligning the VLA's answer “$\mathsf{right/left/up/down/front/back/grasped}$” and predicted actions with the ground-truth.
    % At inference, we use the same data flow to derive the predicted actions. 
    % InSpire is compatible with existing autoregressive VLAs.
    % By employing this visual question answering task as the bridging “language” and performing spatial reasoning and action prediction at the same time, InSpire endows VLAs with spatial reasoning capabilities without employing extra data or interacting with other large models.
    }
\label{fig.method}
\end{figure*}

Our solution the question is \textbf{\underline{In}trinsic \underline{Sp}at\underline{i}al \underline{Re}asoning (InSpire)}, a simple yet effective approach that boosts VLAs' spatial reasoning capabilities to mitigate the adverse effects of spurious correlations on their generalization in novel scenarios.
% Built upon the miniVLA\footnote{\url{https://ai.stanford.edu/blog/miniVLA}} architecture, our Spatial-VLA employs a Qwen 2.5 0.5B \cite{yang2024qwen2} backbone and retains the same ViT for visual encoding as OpenVLA \cite{kim2024openvla}, effectively reducing OpenVLA's footprint from 7B to 1B parameters, as illustrated in Fig. 2.
As presented in Fig. \ref{fig.method}, our proposed InSpire approach redirects the model's attention from spurious factors to task-relevant ones by simply appending the question “$\mathsf{In\,\, which \,\, direction\,\, is\,\, the\,\, [object]\,\, relative\,\, to\,\, the\,\, robot ?}$” before the language instruction and aligning the VLA's generated answer “$\mathsf{right/left/up/down/front/back/grasped}$” and predicted actions with the ground-truth.
% Spatial-VLA takes input the question “\textit{In which direction is the [object] relative to the robot}” together with the language instruction to  
By using this spatial reasoning VQA task as the bridging “language” between observations and actions, InSpire equips VLAs with the ability to understand and reason about spatial relationships without the need to collect auxiliary training data or interact with other large models. Notably, the InSpire approach is fully compatible with existing autoregressive VLAs and can be seamlessly integrated as a \textit{\textbf{plugin}} to enhance their performance. 
We conduct extensive evaluations using both simulation benchmarks (LIBERO \cite{liu2023libero} and CALVIN \cite{mees2022calvin}) and real-world tasks, the achieved results on two state-of-the-art VLAs---{miniVLA-VQ}\cite{miniVLA_blog} (a lightweight version of OpenVLA\cite{kim2024openvla}) and $\pi_0${-FAST} \cite{pertsch2025fast} (an upgraded version of $\pi_0$\cite{black2024pi0}), prove InSpire's effectiveness and flexibility.

To summarize, our contributions in this work are threefold.
\begin{itemize}[itemsep=1ex, topsep=0pt, parsep=0pt] 
    \item We propose InSpire, a novel approach designed to mitigate the negative impact of spurious correlations on the generalization performance of VLAs.
    % by stimulating the inherent spatial reasoning ability of VLMs. 
    \item Without employing extra data or interacting with other large models, InSpire endows VLAs with spatial reasoning capabilities in a plug-and-play manner.
    \item Comprehensive evaluations in both simulation and real-world environments demonstrate the effectiveness and flexibility of the proposed InSpire approach.
    
\end{itemize}

\section{Proposed Approach}
% In this section, we incorporate spatial reasoning into our method as the additional thinking and a way to explicitly extract task-relevant information. 
% We begin by formally describing the learning process of the robot policy.
% In this section, we elaborate on our InSpire methid.

\subsection{\textbf{Preliminaries and Problem Statement}}
In the standard supervised or imitation learning framework, we consider an expert demonstration dataset $D=\{(o_i,l_i,a_i)\}_{i=1}^N$, where $o_i\in\mathcal{O}$ denotes an observation, $l_i\in\mathcal{L}$ a language instruction, and $a_i\in\mathcal{A}\subset\mathbb{R}^m$ an action. Each datapoint is sampled from the training distribution $p_{train}(o,l,a)=p(a|o,l)p_{train}(o,l)$, where the ground-truth expert policy $p(a|o,l)$ is independent of the specific choice of training distribution. The aim is to learn an action policy $\pi_\theta(a|o,l)$ that approximates $p(a|o,l)$. In this work, $\pi_\theta(a|o,l)$ are autoregressive VLAs, typically expressive neural networks like transformers pretrained from VLMs, which directly map visual observations $o$ and language instructions $l$ to action tokens $a$ in an autoregressive manner \cite{kim2024openvla,brohan2022rt1,brohan2023rt2,pertsch2025fast}.

Following \cite{higgins2017darla}, we assume that the combined input $[o,l]$ is generated from underlying ``observation factors''. These factors are categorized into task-relevant factors $u \in \mathcal{U}$, which causally determine the action, and task-irrelevant factors $v \in \mathcal{V}$, which have no causal effect on the action. Consequently, the expert policy depends only on task-relevant factors, $p(a|o,l)=p(a|u)$, implying that in the true data-generating process, actions $a$ are independent of task-irrelevant factors $v$ (i.e., $p(a,v)=p(a)p(v)$). Spurious correlations arise when $a$ and $v$ become statistically dependent within the training distribution, such that $p_{train}(a,v) \neq p_{train}(a)p_{train}(v)$. This often occurs if the task-relevant factors $u$ and task-irrelevant factors $v$ are themselves correlated in the training data ($p_{train}(u,v) \neq p_{train}(u)p_{train}(v)$). Given the causal relationship $u \to a$, a correlation between $u$ and $v$ under $p_{train}$ can induce a non-causal statistical association between $v$ and $a$. Learned policies relying on such spurious correlations exhibit poor generalization to distributions beyond $p_{train}$, leading to unreliable performance in novel scenarios.

\subsection{\textbf{Intrinsic Spatial Reasoning}}

We hypothesize that models resort to spurious correlations when task-relevant factors are not explicit in the input. Such latent factors are difficult to learn, as neural networks preferentially learn simpler patterns first \cite{arpit2017closer}. For instance, in a task such as ``put the bowl in the drawer of the cabinet'' (see Fig. \ref{fig.method}), the small size of the bowl might lead the model to infer actions from irrelevant distractors or background elements rather than the object of interest. The central insight behind our method is to enable the model to first extract salient, task-relevant information from observations—a simpler pattern to discern—and then utilize this information as an additional input for action generation. 

\subsubsection{\textbf{Method Overview}} Formally, we denote $u'=f(u)$ as an extracted representation that processes and summarizes high-level information from the task-relevant factors $u$. The goal is to learn two policies: an extraction policy $\pi_{u'}:\mathcal{O}\times\mathcal{L}\rightarrow \mathcal{U}'$ which maps observation and task instruction to the extracted task-relevant representations, and an action policy $\pi_{\theta}:\mathcal{O}\times\mathcal{L}\times \mathcal{U}'\rightarrow \mathcal{A}$ which maps observation, task instruction, and the extracted task-relevant representations to actions. Consequently, the model outputs actions via a two-step process instead of a single step:
\begin{align*}
    \textcolor{red}{u'}&=\pi_{u'}(o,l),\\
    a& =\pi_\theta(\textcolor{red}{u'},o,l).
\end{align*}
The extracted representation $u'$ is designed to be more readily learnable, thereby guiding the model away from spurious correlations present in the original observations $o$.

\subsubsection{\textbf{Modeling Task-relevant Factors via Spatial Reasoning VQA}}
To establish a reliable representation $u'$ of task-relevant factors, we use the extensive, text-based world knowledge inherent in VLMs—the models from which contemporary VLAs are pretrained. Exploiting text as this representation $u'$, we propose Intrinsic Spatial Reasoning (InSpire) that performs explicit spatial reasoning regarding the robot and objects of interest prior to the inference of subsequent actions.

\begin{figure}[t]
% \begin{wrapfigure}{l}{0.4\textwidth}
	\centering
% \vspace{-0.3cm}
\includegraphics[width=0.75\linewidth]{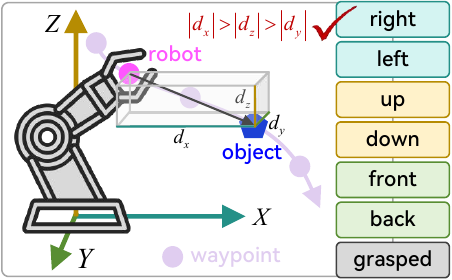} 
	\caption{\textbf{Automated rule-based object direction labeling.} At each waypoint of a trajectory, the 3D locations of the robot's gripper and target objects are obtained from the simulation environment or recorded positions where the robot interacts with objects in the real-world environment. These locations are used in a {rule-based} strategy to {automatically} compute the object's direction.}
\label{fig.labels}
\vspace{-0.3cm}
\end{figure}

As shown in Fig. \ref{fig.method}, we employ a VLA that dually functions as an extraction policy $\pi_{u'}$ and an action policy $\pi_{\theta}$. Given an initial language instruction $l$, InSpire first introduces a textual question $q$ to probe the spatial relationships between objects mentioned in $l$ and the robot, e.g., “$\mathsf{In\,\, which \,\, direction\,\, is\,\, the\,\, [object]\,\, relative\,\, to\,\, the\,\, robot}?$”. Object names are identified within $l$ using the natural language toolkit \cite{loper2002nltk}. The VLA, operating as the extraction policy $\pi_{u'}$, takes $q$, the current visual observation $o$, and the language instruction $l$ as input to produce a textual answer $g$, which is constrained to a predefined set of valid options of coarse-grained directions, as detailed in Fig. \ref{fig.labels}. 
The combination of the self-generated question and its corresponding answer constitutes an extracted textual representation $u' = [q,g]$. This representation $u'$ is subsequently passed to the same VLA, now serving as the action policy $\pi_{\theta}$, which outputs the final robot actions. Because this spatial reasoning VQA task is specifically formulated to correlate highly with objects of interest and the requisite actions, the resulting textual pair $u'$ provides a distilled, abstract representation of task-relevant factors, simplifies the learning challenge for the action policy, and mitigates the learning of spurious correlations.

% \end{wrapfigure}

\subsubsection{\textbf{Automated Rule-based Object Direction Labeling}} To facilitate model training, ground-truth spatial relationships are integrated into the training datasets. Each training datapoint is thus augmented to $(o_i, a_i, u'_i)$, where $u'_i$ encompasses self-generated questions and their corresponding ground-truth spatial relationship answers. These ground-truth relationships are derived by first determining the positions of the robot's end-effector/gripper and relevant objects, then inferring their relative spatial arrangement, as illustrated in Fig.~\ref{fig.labels}. In simulations, object locations are readily available. In real-world settings, the end-effector's position at the instant of gripper closure/opening is recorded and used as a proxy for the object's position when establishing these relationships. 
Given an end-effector position $[x_i,y_i,z_i]$ and an object position $[x_0,y_0,z_0]$, the position difference is calculated as $\bm{d} = [x_i-x_0, y_i-y_0, z_i-z_0]$. The coarse-grained spatial relationship is then determined by the axis corresponding to the component of $\bm{d}$ with the largest absolute value. For instance, if $|x_i-x_0|$ is the maximum component magnitude in $\bm{d}$, the relationship is classified as “left” or “right” based on the sign of $x_i-x_0$. When the object is grasped, as indicated by gripper closure, it is directly labeled with the text “grasped”. This rule-based strategy, applied from a third-person viewpoint, generates the ground-truth labels used for supervising the VLA's spatial reasoning. During training, the autoregressive loss is also applied to tokens corresponding to spatial reasoning answers, alongside the primary action tokens; this additional loss encourages the model to predict the correct textual descriptions of these spatial relationships. Our experiments show that this relatively coarse-grained spatial reasoning is nonetheless effective for guiding the action generation process (see Section \ref{sect.abl}).

\section{Experiments}
In this section, we conduct extensive experiments to answer the following questions:
\begin{mdframed}
[skipabove=4pt, innertopmargin=4pt, innerbottommargin=4pt]
\textit{1) Can InSpire enhance VLAs on tasks from both simulation and real-world environments? } \\
\textit{2) Can InSpire surpass other reasoning-based methods in improving VLAs?}\\
\textit{3) What is the impact of various design decisions on InSpire's performance?}\\
\textit{4) How does InSpire help resolve spurious correlations?}
\end{mdframed}
% \vspace{-5pt}
% For detailed information on determining the relative positions of task-specific objects and the robot in each observation within both simulated and real-world environments, please refer to DDD.
We answer the first question in Sections \ref{sect.simul} and \ref{sect.realworld},  the second question in Section \ref{app.sota}, the third question in Section \ref{sect.abl}, and the fourth question in Section \ref{sect.further}. 
% The baseline VLAs and the evaluation tasks utilized in simulation and real-world environments are described in detail in \textbf{Appendix} \ref{app.baseline}. 
% Besides, we compare our InSpire with state-of-the-art reasoning-based VLAs in \textbf{Appendix} \ref{app.sota}.

% More details about the baseline VLAs, evaluation tasks and hyperparameters used in our simulation and real-world experiments are presented in \textbf{Appendix} \ref{app.setup}. Additional experimental results are provided in \textbf{Appendix} \ref{app.addition}.

\subsection{\textbf{Simulation Experiments}} \label{sect.simul}
We perform simulation experiments using LIBERO \cite{liu2023libero} and CALVIN \cite{mees2022calvin} benchmarks. We train the model with random 3 seeds, and conduct 100 trials per task at inference. 

\subsubsection{\textbf{LIBERO Evaluation}} 
% LIBERO reflects real-world performance, enabling reliable assessment of VLAs. 
We use manipulation tasks from the five datasets—LIBERO-90, LIBERO-Spatial, LIBERO-Object, LIBERO-Goal and LIBERO-Long/10. The VLA model is jointly trained on the 90 tasks in the LIBERO-90 dataset, with each task consisting of 50 demonstrations. To comprehensively assess InSpire’s capabilities, we evaluate the model on both seen tasks from LIBERO-90 and unseen tasks from the other four datasets. 
We perform evaluations using two state-of-the-art models: \textbf{miniVLA-VQ}\cite{miniVLA_blog} and \textbf{$\pi_0$-FAST} \cite{pertsch2025fast}. 
% Rather than using the vanilla miniVLA model, we employ miniVLA-VQ, an enhanced variant of miniVLA that integrates a vector quantization-based action chunking strategy.
In our experiments,  miniVLA-VQ is pretrained from scratch, whereas $\pi_0$-FAST undergoes full-finetuning on the LIBERO-90 dataset.
To ensure fair comparisons, both models adopt the same hyperparameters (e.g., learning rate = 1e-5 and training step = 50000) as those employed for InSpire. 
% In addition to comparing InSpire with the two baselines, we evaluate it against the 7B-parameter OpenVLA model \cite{kim2024openvla}, the vanilla miniVLA and its two other variants miniVLA-VQ-history and miniVLA-VQ-wrist presented in the miniVLA project homepage\footnote{\url{https://ai.stanford.edu/blog/miniVLA}\label{miniVLA}}. 
Since the LIBERO-90 pretrained OpenVLA model has not been released, we borrow the reported results on seen tasks from the project homepage. 
% Unlike prior works \cite{zhao2025cot,michal2024robotic} that artificially create unseen tasks for each dataset in the simulation environment, we sample novel tasks from publicly available out-of-distribution datasets, ensuring result reproducibility and fair comparisons.
% To comprehensively evaluate InSpire’s capabilities, we assess its performance on both seen and unseen tasks. Each simulation task is accompanied by 50 demonstrations. 

\begin{table}[t]
\centering
\tabcolsep 0.02in
\footnotesize 
\caption{\textbf{LIBERO Performance}. $^*$ models borrowed from \cite{miniVLA_blog}. $^\dagger$ our models pretrained from scratch on LIBERO-90.  $^\ddagger$ our models finetuned on LIBERO-90.}
\resizebox{0.49\textwidth}{!}{
\begin{tabular}{l|c|ccccc}
\hline
% \fontsize{6}{10}
\multirow[c]{2}{*}{\textbf{Model}}& \multicolumn{1}{c|}{\textbf{Seen}} & \multicolumn{5}{c}{\textbf{Unseen}}  \\ 
% \cline{2-7}
  & LIBERO-90 & -Spatial & -Object & -Goal & -Long & \textbf{Average} \\ 
\hline
OpenVLA \cite{kim2024openvla} &61.4 &- &- &- &- &- \\ 
miniVLA$^*$ \cite{miniVLA_blog}  & 62.0 &0 &0 &0 &1 & 0.25 \\ 
miniVLA-VQ$^*$ \cite{miniVLA_blog} & 77.0 &0 &0  &3 &1 & 1\\ 
miniVLA-hist.$^*$\cite{miniVLA_blog} & 82.0 & 0&1 &2 &7 & 2.5\\
miniVLA-wrist$^*$\cite{miniVLA_blog} & 82.1& 0& 0&0 &0 &0 \\
SpatialVLA$^\ddagger$\cite{zhang2024spatialvla} & 46.2$_{\pm1.7}$ & 0 & 0 & 0.67 &1.33 &0.5$_{\pm0.5}$ \\
\hline 
miniVLA-VQ$^\dagger$ \cite{miniVLA_blog}  &83.3$_{\pm1.2}$ &{0.7} & 0 &5.7 &8.0 &3.6$_{\pm0.4}$ \\ 
\rowcolor{green!10} \textbf{+InSpire} (\textit{Ours}) &\textbf{89.5}$_{\pm1.5}$  &{\textbf{13.7}}  &{\textbf{20.1}} &\textbf{12.7}  &{8.0} &\textbf{13.6}$_{\pm3.9}$\\ 
% \rowcolor{green!10} \quad \quad  {\textit{relat.}} $\Delta$  & \textcolor{red}{+{8.3}\%} &\textcolor{red}{+{1800}\%} &- &\textcolor{red}{+{123}\%} &\textcolor{red}{+{0}\%} &\textcolor{red}{+{272}\%} \\ 
\rowcolor{green!10} \quad \quad  \quad $\Delta$  & \textcolor{red}{+{6.2}} &\textcolor{red}{+{13.0}} &\textcolor{red}{+{20.1}}  &\textcolor{red}{+{7.0}} &\textcolor{black}{+{0}} &\textcolor{red}{+{10.0}} \\ 
\hline
$\pi_0$-FAST$^\ddagger$ \cite{pertsch2025fast} & 83.1$_{\pm1.0}$ & 5.7 & 6.0 &6.0 &5.0 &5.7$_{\pm0.8}$ \\ 
\rowcolor{green!10} \textbf{+InSpire} (\textit{Ours}) & \textbf{84.1}$_{\pm1.0}$ & \textbf{7.3}  &\textbf{15.7} & 6.0  &5.0  & \textbf{8.5}$_{\pm1.3}$ \\ 
\rowcolor{green!10} \quad \quad \quad $\Delta$ & \textcolor{red}{+{1.0}} & \textcolor{red}{+{1.6}} & \textcolor{red}{+{9.7}}& \textcolor{black}{+0} & \textcolor{black}{+0} & \textcolor{red}{+{2.8}} \\ 
\hline
\end{tabular}}
\label{tab:LIBEROperformance}
\end{table}

% \textbf{Results and Analysis}. 
% Table \ref{tab:LIBEROperformance} reports the performance of the two state-of-the-art autoregressive models {miniVLA-VQ} and $\pi_0$-FAST integrated w/ or w/o our InSpire approach on the LIBERO benchmark.  
We have the following key observations from Table \ref{tab:LIBEROperformance}.
% \textbf{1)} The two baseline VLAs as well as other comparison models exhibit poor generalization performance on all four datasets of unseen tasks, indicating that these models fail to capture true causal relationships between observations and actions, leading to poor generalization beyond the distribution of their training data.
\textbf{1)} Our InSpire achieves the best or comparable performance compared to the two strong baseline VLAs on both seen and unseen tasks. Remarkably, InSpire enhances the absolute success rates of miniVLA-VQ by \textbf{6.2\%} and \textbf{10\%} on seen and unseen tasks, respectively. 
% This underscores InSpire's effectiveness and flexibility in tackling the adverse effects of the spurious correlation issue in VLAs, thereby enabling them to produce more accurate and robust robot actions. 
\textbf{2)} The established performance gains of InSpire on $\pi_0$-FAST are not as significant as that on miniVLA-VQ. The primary reason could be that we train miniVLA-VQ from scratch but finetune the pretrained $\pi_0$-FAST. 
With substantially less data than $\pi_0$-FAST's pretraining dataset, LIBERO-90 limits the ability of InSpire to enhance $\pi_0$-FAST's spatial reasoning capability.
\textbf{3)} InSpire yields significantly better performance on LIBERO-Spatial and LIBERO-Object tasks compared to LIBERO-Goal and LIBERO-Long tasks, demonstrating its strength in handling spatial relationships and object interactions. 
% The results on LIBERO-Long tasks also reveal InSpire's inability to improve VLAs' high-level task planning capacity.
% or subtask decomposition \cite{michal2024robotic,myers2024policy,zhao2025cot}.
\textbf{4)} SpatialVLA, and other VLAs, exhibit poor generalization performance on tasks from the four unseen datasets, indicating these models' inability to capture true causal relationships between observations and actions, leading to poor generalization beyond the distribution of their training data.
% In summary, our InSpire approach can function as a plugin to effectively enhance the VLA's generalization on both seen and unseen tasks.

\subsubsection{\textbf{CALVIN Evaluation}} 
CALVIN \cite{mees2022calvin} is a challenging benchmark for long-horizon, language-conditioned robot manipulation. It features 34 tasks with unconstrained language instructions, performed by a Franka Panda robot in an interactive desk environment. The training dataset contains over 20k expert trajectories paired with language instruction labels 
We follow \cite{mees2022calvin,wuunleashing} to train the model to predict delta XYZ positions and delta Euler angles for arm actions and binary gripper actions. We apply our InSpire approach to two baseline models, miniVLA and miniVLA-VQ. 
% The learning rate is 1e-5, and the batch size is set to 128.
We use the ABC$\xrightarrow{}$D evaluation protocol, where A, B, C, and D stand for four different environments. 
We train the model on environment ABC with three random seeds and evaluate it on a distinct test environment D. The evaluation consists of 500 randomly generated task sequences, each comprising five subtasks. The model attempts the subtasks sequentially, proceeding to the next only upon success. We report the number of successfully completed subtasks in a row.

\begin{figure}[t]
	\centering
	\includegraphics[width=0.9\linewidth]{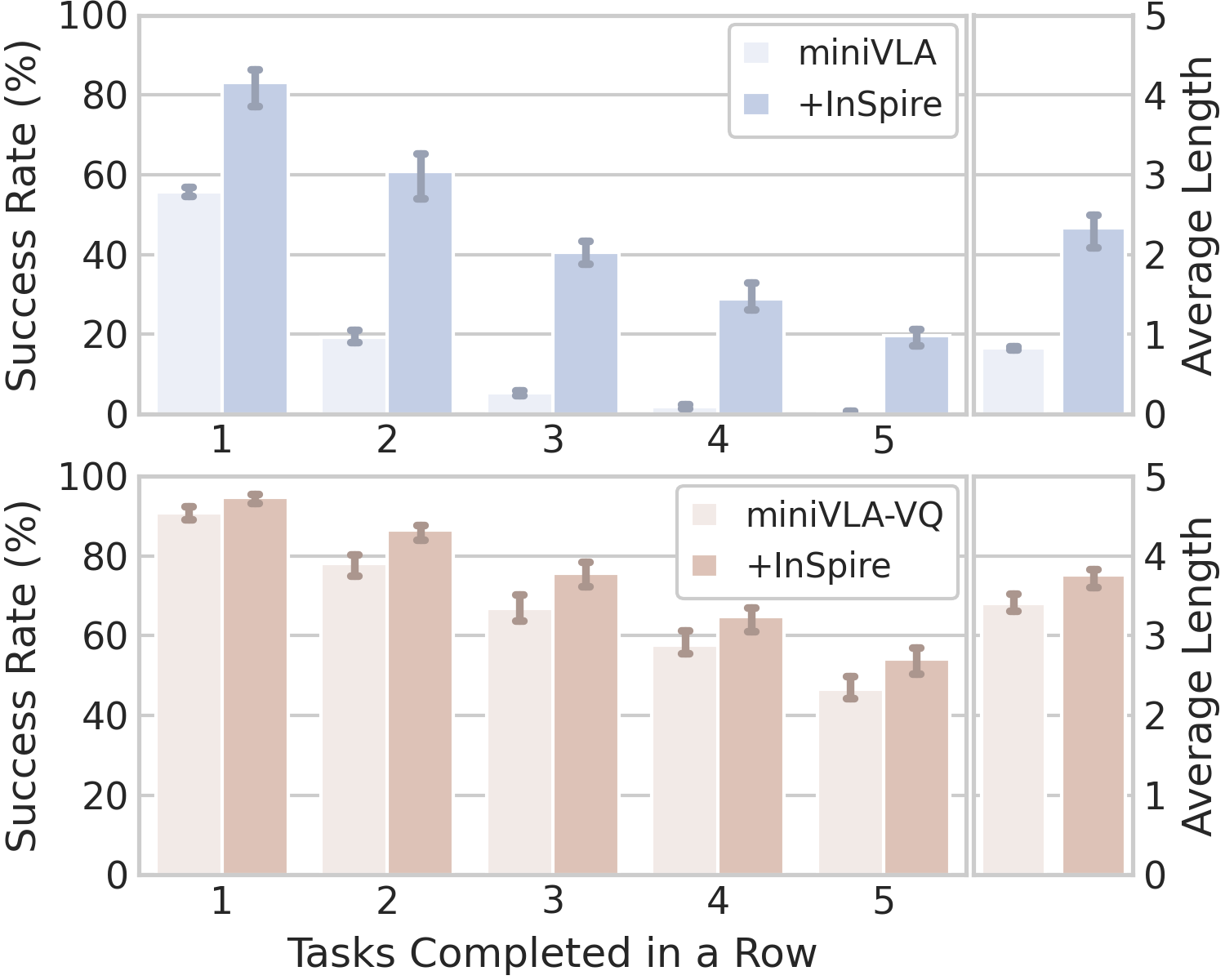} 
	\caption{\textbf{CALVIN Performance}. } 
    \label{fig.Calvin}
    % \label{fig.posi}
\end{figure}

Quantitative results are shown in Fig.\ref{fig.Calvin}.
% As can be observed, our InSpire approach outperforms all the baseline methods on sequentially completing 1, 2, 3, 4, and 5 tasks in a row. The average length in the last column, computed by averaging
% the number of completed tasks in a row of 5 in all the evaluated sequences, shows the long-horizon capability in a comprehensive way. InSpire outperforms all the comparing baseline methods. 
% This demonstrates the superiority of our approach on tackling the adverse effects of spurious correlations, thereby enabling them to produce more accurate and robust robot actions. 
As observed, our InSpire consistently outperforms all baseline methods in completing 1 to 5 sequential tasks. This strong long-horizon capability is further highlighted by the average number of consecutively completed tasks, where InSpire again surpasses all competitors. This superior performance stems from our method's effectiveness in mitigating spurious correlations, leading to more accurate and robust robot actions.

% \textbf{Comparison with State-of-the-arts}. TBD

\subsection{\textbf{Real-world Experiments}} \label{sect.realworld}
% The advantages of our InSpire are clearly showcased through simulation experiments. Here, we undertake extensive experiments on real-world tasks to evaluate InSpire's practical effectiveness.
% The advantages of our InSpire approach are clearly demonstrated through simulation experiments. Here, we conduct extensive experiments on real-world tasks to assess InSpire’s practical effectiveness.

\begin{figure*}[t] 
	\centering
	\includegraphics[width=0.75\linewidth]{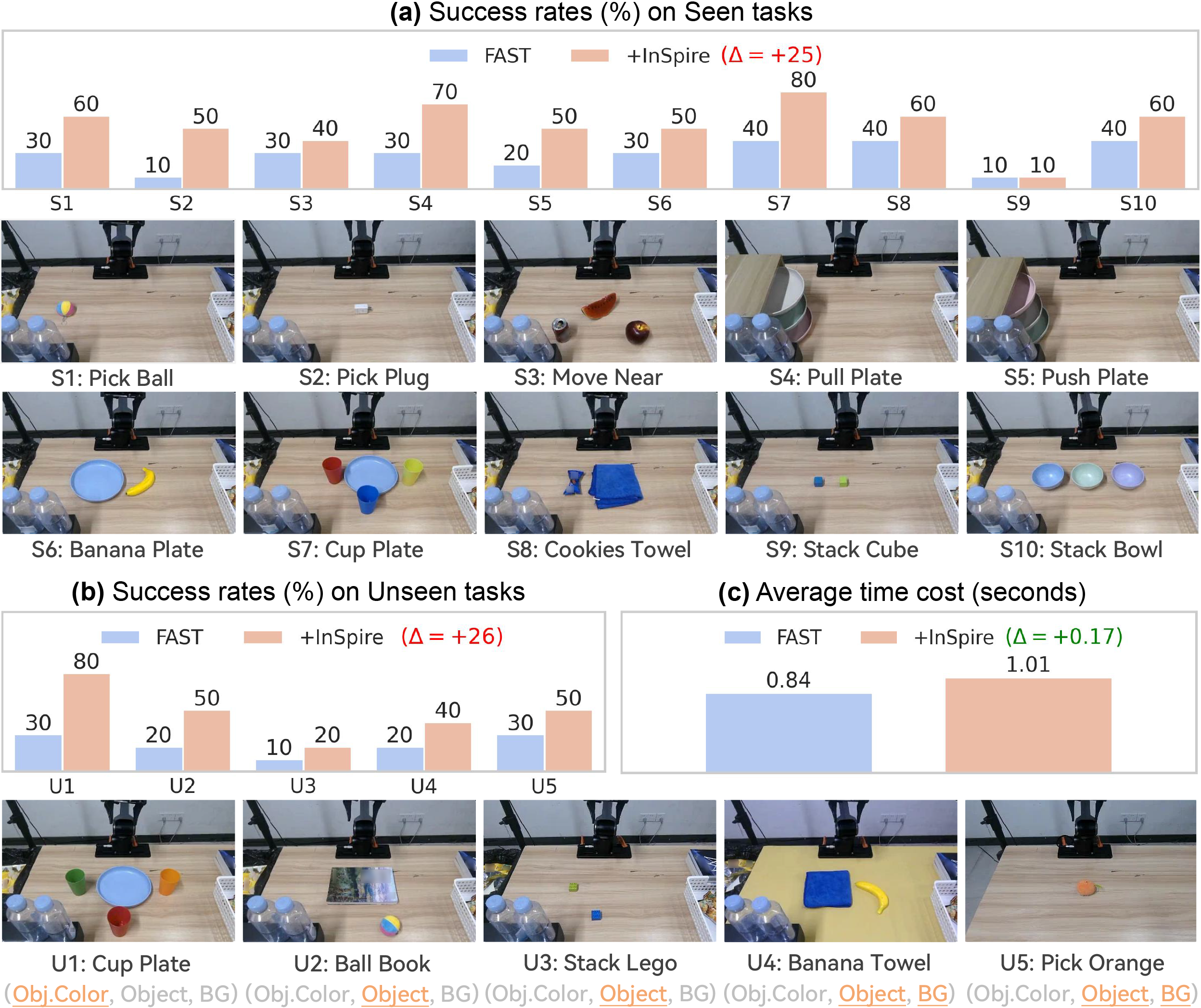} 
	\caption{\textbf{Real-world Performance}. \textbf{(a)(b)} Success rates (\%) of the state-of-the-art model $\pi_0$-FAST \cite{pertsch2025fast} integrated w/ or w/o InSpire on seen and unseen real-world manipulation tasks. \textbf{(c)} Average time cost per step (in seconds) over all seen and unseen tasks. $\Delta$: absolute improvement.}
    %  Detailed language instructions are proved in \textbf{Appendix} \ref{app.task}.
\label{fig.realworld}
\end{figure*}

\begin{table}[t]
\centering
\small
\tabcolsep 0.05in
\caption{\textbf{Comparison with state-of-the-art}. $^\dagger$ independently pretrained from scratch using the four LIBERO datasets. $^\ddagger$ jointly finetuned on the four LIBERO datasets. The results of SpatialVLA-4B are obtained from its original paper \cite{zhang2024spatialvla}, while the results of other rivals are referred from \cite{zhao2025cot}.}
\resizebox{0.48\textwidth}{!}{
\begin{tabular}{l|ccccc}
\hline
\textbf{Model} & {-Spatial} & {-Object} & {-Goal} & {-Long}  & \textbf{Average} \\ \hline
Diffusion Policy$^\dagger$ \cite{chi2023diffusion}  & 78.3 & 92.5 & 68.3 & 50.5 & 72.4$_{\pm0.7}$\\
Octo$^\ddagger$ \cite{team2024octo}  & 78.9 & 85.7 & 84.6 & 51.1& 75.1$_{\pm0.6}$ \\ 
OpenVLA$^\ddagger$ \cite{kim2024openvla}& 84.7 & 88.4 & 79.2 & 53.7 & 76.5$_{\pm0.6}$ \\ 
\hline
SpatialVLA-4B$^\ddagger$ \cite{zhang2024spatialvla} & 88.2 & 89.9 & {78.6} & {55.5} & 78.1$_{\pm0.7}$\\ 
CoT-VLA-7B$^\ddagger$ \cite{zhao2025cot} & {87.5} & 91.6 & {87.6} & {69.0} & 83.9$_{\pm0.6}$\\ 
\rowcolor{green!10} \textbf{InspireVLA-1B} (\textit{Ours})$^\ddagger$ & \textbf{90.7} & \textbf{94.3} & \textbf{88.3} & \textbf{73.3} & \textbf{86.7}$_{\pm1.2}$ \\
\hline
% OTTER$^\dagger$  &  -& 84.0$_{\pm1.0}$ & 89.0$_{\pm1.2}$ & {82.0$_{\pm1.0}$} & - \\ 
% % \rowcolor{green!10} \textbf{InspireVLA} (Ours)  &  &  &  &  & \\
% \rowcolor{green!10} \textbf{InspireVLA} (\textit{Ours})$^\dagger$ &  &  &  &  & \\
% \hline
\end{tabular}}
\label{tab:sota}
\end{table}

% \textbf{Experimental Setup}. 
% Due to the difficulty in collecting a large corpus of real-world behavioral data, it is challenging to train a model from scratch or finetune a pretrained model that hasn't been exposed to real-world manipulation data (e.g., the miniVLA-VQ model).
% In alignment with the simulation experiments, 
Due to the difficulty in collecting huge pools of real-world behavioral demonstrations, it is challenging to train a model that displays generalization and robustness on varied real-world tasks, whether starting from scratch or finetuning a pretrained model lacking exposure to substantial real-world manipulation data.
Hence, we employ $\pi_0$-FAST, the current best-performing open-source VLA, as the baseline model for real-world evaluations.
We carefully design 10 seen tasks and 5 unseen tasks that focus on evaluating the VLA’s performance along multiple dimensions: spatial reasoning, interacting with novel objects and scenes, and following unknown instructions.
% as shown in \textbf{Appendix} {\ref{app.task}}.
% We design 10 challenging manipulation tasks as seen tasks. We modify their language instructions as well as corresponding objects to construct 5 unseen tasks, as shown in \textbf{Appendix} {\ref{app.task}}. 
% In alignment with the simulation experiments, 
The $\pi_0$-FAST model is full-finetuned using the 10 seen tasks, each consisting of training 10 training trajectories. 
During inference, we conduct 10 trials per task and randomize the configurations and orientations of task-specific objects for each trial. We use an AGILEX PiPER 6DOF robot arm. Other hyperparameters are consistent with those used in the simulation experiments.

% \textbf{Results and Analysis}. 
Fig. \ref{fig.realworld} reports the success rates as well as time costs of $\pi_0$-FAST \cite{pertsch2025fast} integrated w/ or w/o our InSpire on 10 seen and 5 unseen real-world tasks. 
% For each task, we record the time cost (in seconds) from the start of the inference process to the successful completion of the task. 
From the obtained results in the figure, we have several key observations.
\textbf{1)} InSpire consistently enhances the success rates of the strong baseline VLA across 10 seen and 5 unseen tasks, achieving an average enhancement of \textbf{25}\% and \textbf{26}\%, respectively.
\textbf{2)} InSpire achieves notable performance gains and exhibits robustness against variations in object color, objects, and backgrounds in unseen tasks. Particularly, InSpire increases the relative success rate of the baseline by \textbf{100}\% on 4/5 of the unseen tasks.
% For example, in the challenging "Stack Cube" task, PCD improves the success rate from 0.05 to 0.10, showing its strong adaptability and robustness across diverse scenarios.
\textbf{3)} InSpire incurs an additional time cost of 0.18 seconds per step on average compared to the baseline VLA. The additional computational overhead is primarily attributed to the spatial reasoning VQA task introducing more tokens into the model's action generation process. 
However, considering the notable improvement in success rates, this trade-off is acceptable in a wide range of real-world robotic applications.
\textbf{4)} The backgrounds of the 15 real-world tasks are significantly more complex than those of the simulation tasks from the LIBERO environment, with numerous distracting elements like plastic bottles, baskets, and trash cans. Nevertheless, our InSpire demonstrates outstanding performance, consistent with the results observed in simulation experiments.
% It is worth mentioning that except for the amount of learning steps, we use consistent InSpire hyperparameters across tasks in both the simulation and real-world environments, without carefully tuning task-specific PCD hyperparameters for each individual task, leaving room for potential performance improvements through task-specific parameter tuning.
% To summarize, the results achieved by InSpire on both seen and unseen real-world tasks underscore its effectiveness and adaptability in real-world robotic systems.

\begin{table*}[t]
\centering
\tabcolsep 0.05in
\footnotesize 
\caption{{Ablation of the formulation of the spatial reasoning VQA task}.}
\resizebox{\textwidth}{!}{
\begin{tabular}{l|l|l|c|c}
\hline
\multirow[c]{1}{*}{\textbf{Setting}}& \quad \quad \quad  \quad \quad \quad \quad \quad \quad \, \textbf{Question}& \quad \quad \quad \textbf{Answer}& \multicolumn{1}{c|}{\textbf{Seen}} & \multicolumn{1}{c}{\textbf{Unseen}}  \\ 
\cline{2-5}
\hline
Baseline & \quad \quad \quad  \quad \quad \quad \quad \quad \quad \quad \quad \textcolor{black}{-} & \quad \quad \quad \quad \textcolor{black}{-} &83.3$_{\pm1.2}$ &3.6$_{\pm0.4}$  \\ \hline 
\rowcolor{green!10} 1D  Direct. &\textcolor{black}{\scriptsize{$\mathsf{In\, which \, direction\, is\, the\, [object]\, relative\, to\, the\, robot?}$} } & \textcolor{black}{\scriptsize{$\mathsf{right/left/.../back/grasped}$} }&\textbf{89.5}$_{\pm1.5}$ &{13.6}$_{\pm3.9}$\\ 
3D Direct. & \textcolor{black}{\scriptsize{$\mathsf{In\, which \, direction\, is\, [object]\, relative\, to\, the\, robot? \, x, y, z: }$} }&\textcolor{black}{\scriptsize{$\mathsf{[right,up,front]/.../grasped}$}} & 87.9$_{\pm2.1}$ &\textbf{15.0}$_{\pm1.9}$ \\
Proximity &\textcolor{black}{\scriptsize{$\mathsf{What \, is\, the \, distance\, between\, the \, robot\, and\, [object]?}$} }&\textcolor{black}{\scriptsize{$\mathsf{far/middle/near/grasped}$}}&
88.7$_{\pm1.1}$ &10.3$_{\pm1.0}$\\
3D Locat. & \textcolor{black}{\scriptsize{$\mathsf{What \, is\, the\, accurate\, posi.\, of \, [object]\, relat.\, to\, the\, robot? \,  x, y, z: }$}} & \textcolor{black}{\scriptsize{$\mathsf{[1,-3,4]/.../[2,0,1]}$} } &88.1$_{\pm0.8}$ &10.1$_{\pm2.6}$ \\
Distance&\textcolor{black}{\scriptsize{$\mathsf{What \, is\, the\,  accurate \,distance\, between\, the \, robot\, and\, [object]?}$}} & \textcolor{black}{\scriptsize{$\mathsf{0/1/2/3/.../9}$}} &85.1$_{\pm2.2}$ &6.6$_{\pm0.6}$ \\
\hline
\end{tabular}}
\label{tab:ablation}
\end{table*}

\subsection{\textbf{Comparison with State-of-the-art}} \label{app.sota}
% Our proposed InSpire scheme aims to enhance the spatial reasoning capabilities of VLAs, thereby overcoming the adverse effects of spurious correlations on their generalization and robustness in unseen scenarios.
In this part, we first apply InSpire to the miniVLA-VQ model and pretrain a 1B-parameter VLA on the LIBERO-90 dataset, which we refer to as \textbf{InspireVLA-1B}. 
We then compare our InspireVLA-1B with state-of-the-art reasoning-based VLAs, including SpatialVLA-4B\cite{zhang2024spatialvla}—a 4B-parameter VLA that injects 3D information into the input observations to achieve spatial-aware action prediction, and CoT-VLA-7B \cite{zhao2025cot}—a 7B-parameter VLA that enhances action prediction through CoT reasoning.
% visual chain-of-thought (CoT)
We follow the common setup in \cite{zhang2024spatialvla,zhao2025cot}, leveraging tasks from LIBERO-Spatial, -Object, -Goal and -Long for evaluation. 
Each of SpatialVLA-4B, CoT-VLA-7B and our InspireVLA-1B is jointly fine-tuned on tasks from the four LIBERO datasets. 
We compare the performance on seen tasks from the four datasets and directly borrow the results of SpatialVLA-4B and CoT-VLA-7B from the original papers.
We perform LoRA fine-tuning \cite{hu2022lora} on InSpireVLA-1B using tasks from the four datasets.
% the hyperparameters adopted in our experiments are listed in Table \ref{tab:hyperparameters3}. 
We report the average success rates and std errors over 3 random seeds. 
As in \cite{zhao2025cot}, we also compare InSpire with other three strong baselines: Diffusion Policy\cite{chi2023diffusion}, Octo\cite{team2024octo} and OpenVLA\cite{kim2024openvla}, their results are directly referred from \cite{zhao2025cot}\footnote{Since the LIBERO-pretrained checkpoints of Diffusion Policy, Octo, OpenVLA and CoT-VLA-7B in \cite{zhao2025cot} have not been open-sourced, we are unable to compare the performance of those models on unseen tasks.}. 

% \textbf{Results and Analysis}. 
From the obtained results in Table \ref{tab:sota}, we have the following key observations.
\textbf{1)} Our InspireVLA-1B model consistently outperforms the five VLAs across the four LIBERO datasets. Particularly, InspireVLA-1B improves the two reasoning-based models SpatialVLA-4B and CoT-VLA-1B by \textbf{8.6}\% and \textbf{2.8}\%, with \textbf{4}$\times$ and \textbf{7}$\times$ parameter efficiency, respectively. 
\textbf{2)} The three reasoning-based models (i.e., SpatialVLA-4B, CoT-VLA-7B and our InspireVLA-1B) achieve better performance than  other three competitors, emphasizing the necessity of integrating an intermediate reasoning process into existing VLAs to capture the true causal relationships between observations and actions. 
\textbf{3)} InspireVLA-1B demonstrates remarkable advantages on LIBERO-Spatial and LIBERO-Object, highlighting its effectiveness in transferring knowledge about spatial information and objects. 
% \textbf{4)} InspireVLA-1B

\subsection{\textbf{Ablations on Design Decisions}} \label{sect.abl}
In this section, we conduct ablation studies in the LIBERO environment to explore how the performance of InSpire varies with different design decisions. We employ miniVLA-VQ as the baseline VLA, and conduct 100 trials per task. Our ablation studies use a single fixed seed as in \cite{zhao2025cot,zawalski2024ecot,nakamotosteering}.

\begin{figure}[t]
% \begin{wrapfigure}{l}{0.4\textwidth}
% \setlength{\abovecaptionskip}{0.1cm}  
% \setlength{\belowcaptionskip}{-0.3cm} 
	\centering
	\includegraphics[width=0.7\linewidth]{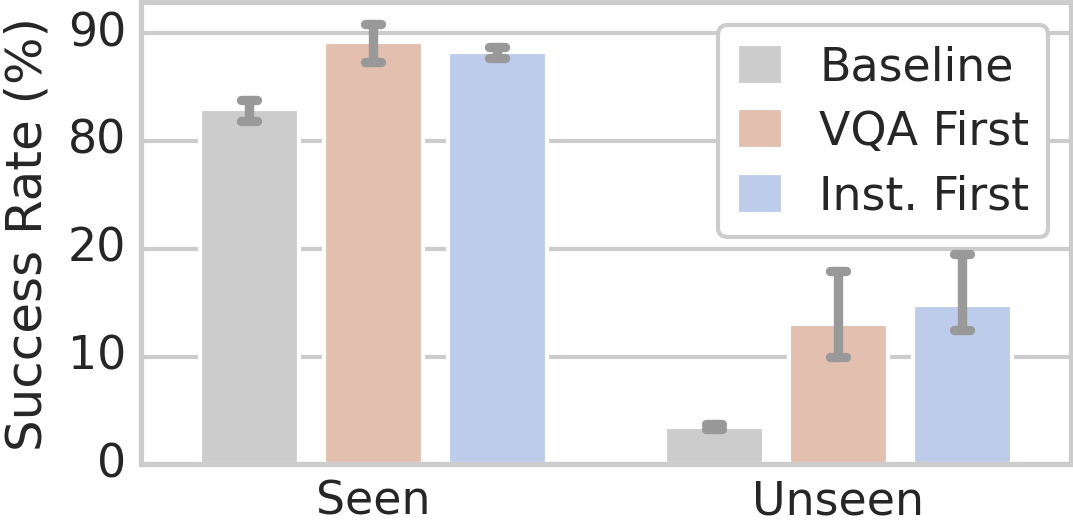} 
	\caption{Ablation study on the insertion position of the spatial reasoning VQA task.} \label{fig.vqa}
    \label{fig.posi}
\end{figure}
% \end{wrapfigure}

% \end{wrapfigure}

\subsubsection{\textbf{VQA Formulations}}
% \textbf{Impacts of Different Kinds of Spatial Information}. 
% Without using extra data or interacting with other large models, 
InSpire leverages a spatial reasoning VQA task as the bridging “language” between observations and actions.
This implies  that the formulation of the VQA task dictates the types of spatial information used by InSpire. Here, we investigate the impacts of different VQA formulations on InSpire's performance.
% It is necessary to answer the following question: What kinds of spatial information does our InSpire need? 
As shown in Table \ref{tab:ablation}, we design five VQA tasks to introduce diverse spatial information, denoted as “1D Direction”, “3D Direction”, “Proximity”, “3D Location” and “Distance”.
The performance of our InSpire approach integrated with each of these VQA tasks is reported in the table. 
We have the following observations from the results in the table.
\textbf{1)} The four VQA formulations consistently outperform the baseline on both seen and unseen tasks.
\textbf{2)} “1D Direction” achieves the best performance among all VQA formulations, surpassing the baseline by large margins.
\textbf{3)} “3D Direction” demonstrates performance close to that of “1D Direction”, with both significantly outperforming the other VQA formulations. 
% One possible reason may be that the direction information, denoted by natural language tokens, .
This suggest that direction prediction-based VQA tasks are more effective in guiding the action generation process.

\subsubsection{\textbf{VQA Insertion Positions}}
% \textbf{Impacts of Different VQA Insertion Positions}. 
In our developed InSpire approach, the tokens for the spatial reasoning VQA task are appended following the input vidsual observation tokens and preceding the language instruction tokens (Fig. \ref{fig.method}).
In this experiment, we investigate the impacts of different insertion positions of those VQA tokens on performance. 
Fig. \ref{fig.posi} presents a quantitative comparison of InSpire’s performance with two design decisions: inserting VQA tokens before the language instruction tokens (“VQA-First”) and placing them after the language instruction tokens (“Instruct-First”).
As seen, “VQA-First” achieves superior performance on seen tasks compared to “Instrut-First”, but lags behind on unseen tasks. Moreover, InSpire substantially boosts the baseline's performance in both settings, highlighting the benefits of the auxiliary VQA task for enhancing spatial reasoning.

% Predicted trajectories of the baseline model miniVLA-VQ integrated w/ or w/o our InSpire on three \textit{unseen} tasks in the LIBERO environment.
\begin{figure*}[t] 
	\centering
	\includegraphics[width=1\linewidth]{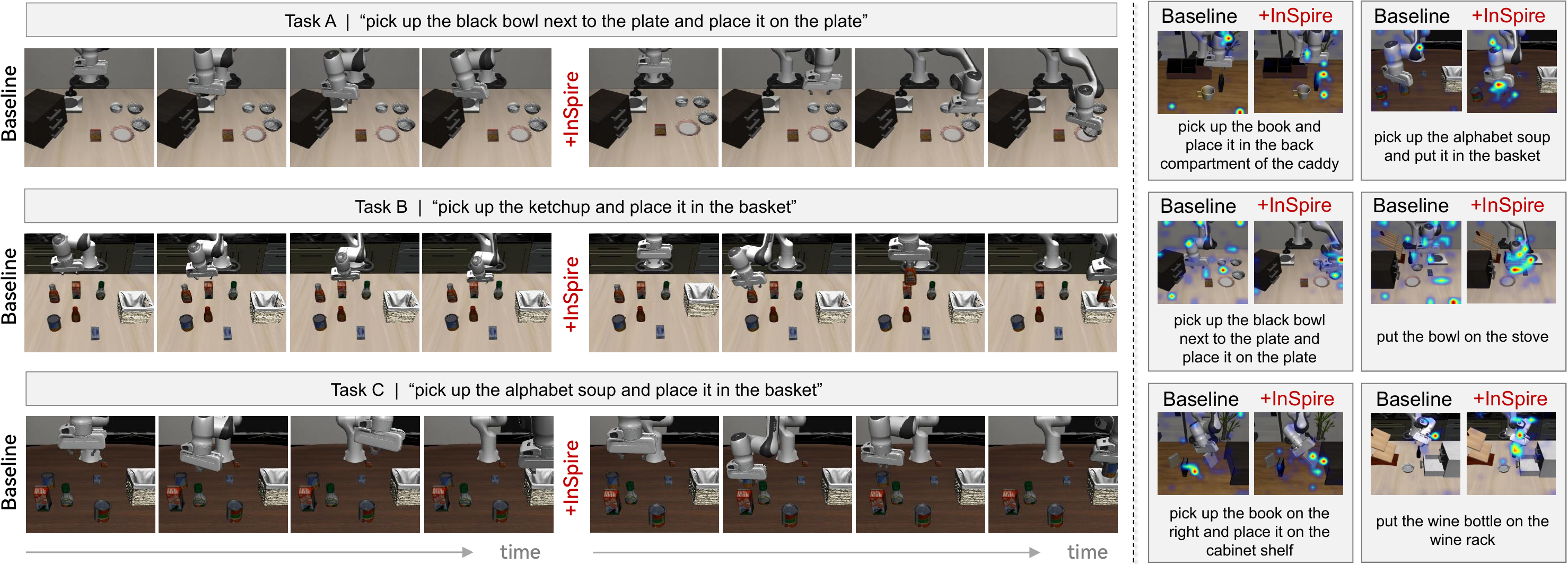} 
	\caption{\textbf{Qualitative Results}. (Left) Action sequences and (Right) attention maps produced by the baseline model (mini-VLA-VQ \cite{miniVLA_blog}) integrated w/ or w/o our proposed InSpire approach. More qualitative results in both simulation and real-world environments are available at: \protect \textcolor{magenta}{\url{https://koorye.github.io/proj/Inspire}}}
\label{fig.demo}
\end{figure*}

\subsection{\textbf{How Does InSpire Help Resolve Spurious Correlations?}} \label{sect.further}
% The core idea of our InSpire approach is to enhance the spatial reasoning capabilities of VLAs to address the side effects of spurious correlations.
In this section, we present qualitative results to explore how InSpire—and spatial reasoning more generally—help mitigate the adverse effects of spurious correlations.
Fig. \ref{fig.demo} illustrates some representative failure modes encountered by the baseline model miniVLA-VQ but successfully addressed by InSpire. 
The results in the figure reveal several strengths of our InSpire approach:
% None of the three manipulation tasks were exposed during training.

\subsubsection{\textbf{Addressing Shortcut Learning of VLAs}}
% VLAs tend to learn spurious correlations between actions and task-irrelevant features of observations. 
As shown in Fig. \ref{fig.demo} (Left) \textbf{Task A}, the baseline VLA, upon observing the “drawer” frequently seen during training, directly approaches it based on shortcut learning without understanding the input language instruction and visual information. In other words, the VLA learns spurious correlations between 
task-irrelevant features of observations and actions, leading to overfitting to seen tasks and poor generalization on unseen tasks. 
Our InSpire approach corrects the shortcut learning exhibited by the baseline, avoiding the neglect of essential elements like language instructions and spatial relations.

\subsubsection{\textbf{Improving Model Robustness to Distractors}}
As illustrated in Fig. \ref{fig.demo} (Left) \textbf{Task B}, the baseline VLA struggles to identify task-specific objects in complex scenarios with various distractors, limiting its ability to produce correct actions. 
The primary reason may be that the direct observation-to-action mapping mechanism prevents the learned model from capturing causal relationships between observations and actions, making it sensitive to familiar distractors when predicting actions.
By employing a spatial reasoning VQA task as the bridging “language” between observations and actions,  InSpire equips the VLA with spatial reasoning capability, thereby helping the VLA differentiate task-relevant objects from  distractors (as shown in Fig. \ref{fig.demo} (Right)).

\subsubsection{\textbf{Enabling Continual Action Correction}}
% TBD. Black bowl / white bowl
Continual correction of incorrect actions is an essential skill for both humans and robots.
Yet, as shown in Fig. \ref{fig.demo} (Left) \textbf{Task C}, the baseline VLA fails to correct the incorrect action and proceeds with following actions.  This suggests the learned VLA uses perceptual data for decision-making in a very different way from how humans do. 
Conversely, InSpire assists the model in continuously correcting errors until the task is completed.
This is primarily because InSpire performs spatial reasoning for each visual observation along the trajectory, which enhances the model's awareness of task execution status (e.g., the relative direction of the target object to the robot, the status of the gripper) and assists in correcting incorrect actions.

%===============================================================================

% Although the results are encouraging, two key limitations remain in this work:
% \textbf{1)} Although InSpire has potential for integration with any LLM-based VLAs, its effectiveness in enhancing diffusion policy-based VLAs, such as Octo \cite{team2024octo} and $\pi_0$ \cite{black2024pi0}, remains to be explored.  Future work will focus on exploring strategies to facilitate seamless integration of InSpire into such models.
% \textbf{2)} The relationships and differences in spurious correlations among different VLAs have not been investigated. VLAs pretrained with diverse paradigms and datasets may acquire varied spurious correlations. Conducting further empirical studies on these factors could improve the effectiveness and robustness of our approach in more complex robotic scenarios.

\section{Related Work}

\subsubsection{\textbf{Vision-Language-Action Models (VLAs)}} VLAs, built upon pretrained Vision-Language Models (VLMs) \cite{Karamcheti2024,Zhai2023}, demonstrate strong generalization in robotics \cite{brohan2023rt2, kim2024openvla}. This line of research, initiated by transformer-based frameworks like RT-1 \cite{brohan2022rt1} and advanced by transferring web-scale knowledge in RT-2 \cite{brohan2023rt2}, has scaled to large datasets such as Open X-Embodiment \cite{embodimentcollaboration2024openxembodimentroboticlearning}. Recent architectures leverage powerful foundation models (OpenVLA \cite{Kim2024}), target computational efficiency (Octo \cite{Team2024}), or refine training and control strategies ($\pi_0$\cite{black2024pi0}, $\pi_0$-FAST \cite{pertsch2025fast}).

% liang2023code,zeng2022socratic,sharma2022skill,myers2024policy,belkhale2024rt,
\subsubsection{\textbf{Chain-of-Thought (CoT) Reasoning}} CoT reasoning has been adapted from NLP to high-level robotic task planning \cite{zawalskirobotic,kang2024clip,shi2025hi,li2025hamster,nasiriany2025rt,kim2025fine}. Applications include sub-goal planning via diffusion or video generation \cite{ni2024generate} and reward shaping in reinforcement learning \cite{zhang2024learning}. Models like CoT-VLA \cite{zhao2025cot} and ECoT \cite{zawalski2024ecot} use visual goal prediction and iterative reasoning, respectively, but can be inefficient by depending on complex intermediate computations or external models. The most related work, RT-H \cite{belkhale2024rt}, uses language motion prediction as an intermediate step. In contrast, our InSpire framework employs a spatial reasoning VQA task to explicitly associate objects of interest with corresponding actions, reducing the impact of spurious correlations.

\subsubsection{\textbf{Learning Spatial Reasoning}} 
Leveraging the spatial reasoning of large models is a recent focus in multimodal learning \cite{lu2023mind,chen2024learning,li2023spatialrgpt,wald2020learning3dsemanticscene,hildebrandt2020scenegraphreasoningvisual,liu2023large,li2023situational}. Methods range from using 2D VLMs for spatial understanding without explicit 3D data (SpatialVLM \cite{chen2024spatialvlm}) to integrating 3D scene graphs and depth information (SpatialRGPT \cite{li2023spatialrgpt}). Our work is most related to SpatialVLA \cite{zhang2024spatialvla}, which injects 3D information and uses discretized action grids. However, instead of relying on encoded 3D positions, our approach employs a VQA task as a bridging “language” to enable simultaneous spatial reasoning and action prediction, endowing VLAs with this capability without extra data or models.

\section{Conclusion, Limitations and Future Work}
This work proposes Intrinsic Spatial Reasoning (InSpire) to mitigate the adverse effects of spurious correlations on the action prediction of VLAs. 
By incorporating a spatial reasoning VQA task as the bridging “language” between visual observations and low-level actions, InSpire endows VLAs with spatial reasoning capabilities without employing extra data or interacting with other large models.
Notably, InSpire can be used as a plugin to improve existing autoregressive VLAs.
Comprehensive experiments demonstrate InSpire's effectiveness and flexibility, achieving consistent improvements on two state-of-the-art VLAs on both seen and unseen tasks in simulation and real-world environments. 
% We hope this work provides a perspective on enhancing the generalization and robustness of VLAs for downstream tasks.

Although the results are encouraging, two key limitations remain in this work:
\textbf{1)} Although InSpire has potential for integration with any LLM-based VLAs, its effectiveness in enhancing diffusion policy-based VLAs, such as Octo \cite{team2024octo} and $\pi_0$ \cite{black2024pi0}, remains to be explored.  Future work will focus on exploring strategies to facilitate seamless integration of InSpire into such models.
\textbf{2)} VLAs pretrained with diverse paradigms and datasets likely acquire varied spurious correlations; however, the characters of these differences across models are not yet understood. A deeper empirical study of these factors is a promising direction for bolstering the robustness of our approach in more complex scenarios.
% The relationships and differences in spurious correlations among different VLAs have not been investigated. VLAs pretrained with diverse paradigms and datasets may acquire varied spurious correlations. Conducting further empirical studies on these factors could improve the effectiveness and robustness of our approach in more complex robotic scenarios.


% \section*{ACKNOWLEDGMENT}
% The preferred spelling of the word ÒacknowledgmentÓ in America is without an ÒeÓ after the ÒgÓ. Avoid the stilted expression, ÒOne of us (R. B. G.) thanks . . .Ó  Instead, try ÒR. B. G. thanksÓ. Put sponsor acknowledgments in the unnumbered footnote on the first page.

%%%%%%%%%%%%%%%%%%%%%%%%%%%%%%%%%%%%%%%%%%%%%%%%%%%%%%%%%%%%%%%%%%%%%%%%%%%%%%%%

% References are important to the reader; therefore, each citation must be complete and correct. If at all possible, references should be commonly available publications.

\bibliographystyle{abbrv}
\bibliography{ref}

\end{document}